\documentclass{article}
\usepackage{ijcai26}

\usepackage{times}
\usepackage{soul}
\usepackage{url}
\usepackage[hidelinks]{hyperref}
\usepackage[utf8]{inputenc}
\usepackage[small]{caption}
\usepackage{graphicx}
\usepackage{amsmath}
\usepackage{amsthm}
\usepackage{amssymb}
\usepackage{booktabs}
\usepackage{algorithm}
\usepackage{algorithmic}
\usepackage{multirow}
\usepackage{booktabs}
\usepackage[switch]{lineno}
\usepackage{graphicx}
\usepackage{xcolor}
\usepackage{xcolor,colortbl} 
\usepackage{subcaption}
\usepackage{tabularx}

\title{Evaluating Explanation-Driven Vision–Language Reasoning via \\Generation Order Interventions}

\author{
Siting Liang$^{1,2}$
\and
Luca Rippe$^2$\and
Omar Adjali$^1$\And
Daniel Sonntag$^{1,2}$\\
\affiliations
$^1$German Research Center for Artificial Intelligence\\
$^2$University of Oldenburg\\
\emails
\{siting.liang, daniel.sonntag, omar.adjali\}@dfki.de,
luca.rippe@uni-oldenburg.de
}

\begin{document}

\maketitle

\begin{abstract}
Natural language explanation generation serves as a key mechanism for exposing and evaluating vision–language reasoning. Prior work on explanation-driven vision–language models predominantly follows a post-hoc (answer-first) paradigm, implicitly suggesting that supervised rationales can reflect underlying reasoning processes. In contrast, modern large vision–language models increasingly exhibit a rationale-first generation tendency, which more closely aligns with structured, stepwise reasoning. In this work, we systematically evaluate whether explanations are causally tied to model predictions within a single generation step under a controlled experimental setup, explicitly eliminating unnecessary chain-of-thought or other intermediate reasoning processes across knowledge-intensive QA, visual entailment, and compositional grounding benchmarks. We find that larger models emerge as a prerequisite for reliably supporting rationale-first reasoning at scale. However, answer-first generation is less prone to format-related errors in structured output. Overall, explanation ordering, model scale and pre-training knowledge, task-specific fine-tuning, and task structure jointly influence both prediction accuracy and reasoning faithfulness.

\end{abstract}

\section{Introduction}
\begin{figure}[t]
    \centering
    \includegraphics[width=0.9\linewidth]{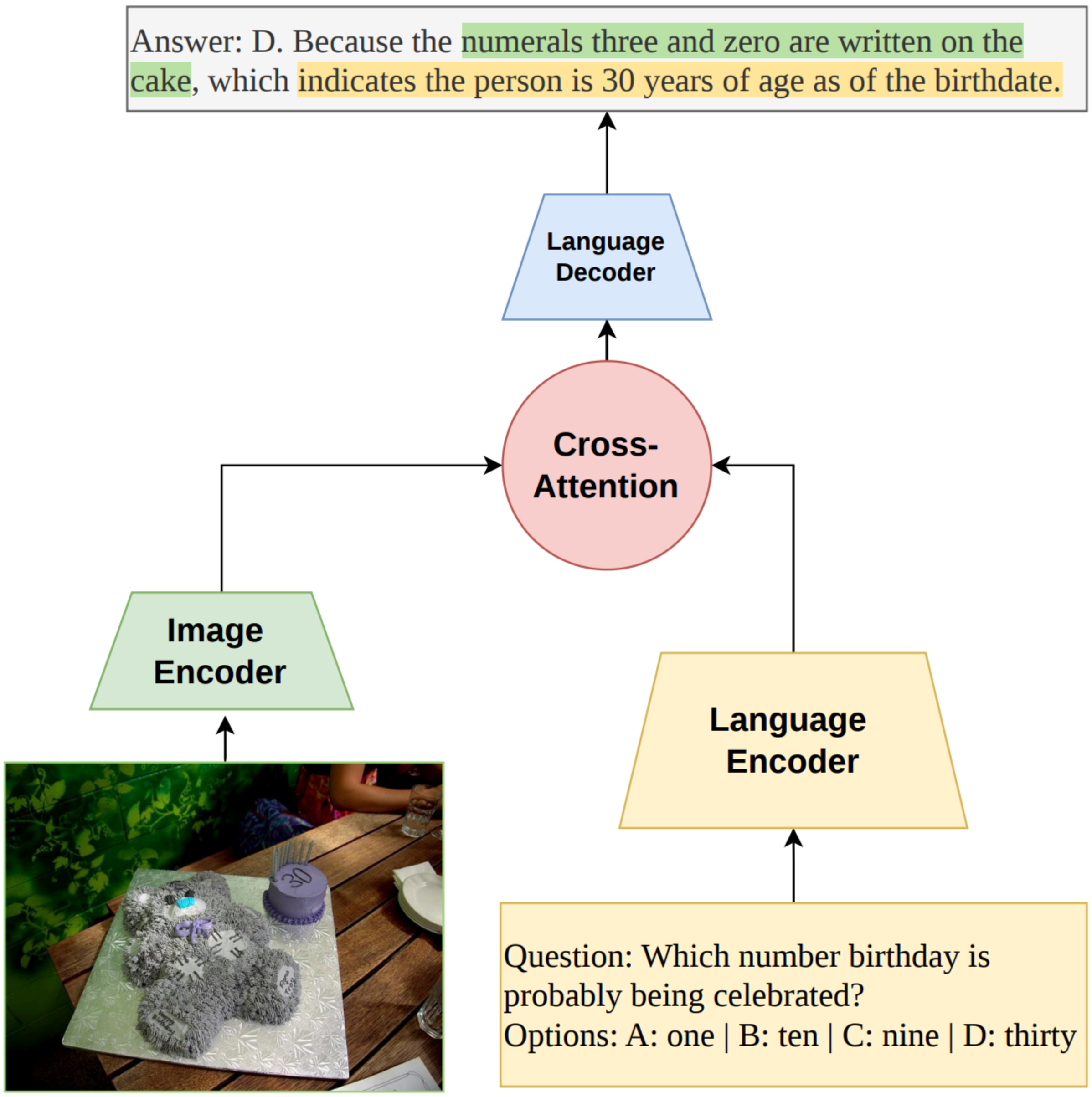}
    \caption{Conventional explanation‑driven vision–language reasoning through post‑hoc rationalization. The model generates an answer first and then produces post‑hoc explanations, primarily grounding the justification in visual evidence, often with limited use of world knowledge.  }
    \label{fig:VLR}
\end{figure}

Previous work on explanation-driven vision-language reasoning (VLR), such as  e-vil\cite{Kayser_2021_ICCV_eug} and nlx-gpt \cite{Sammani_2022_CVPR_nlx_gpt}, constructs training datasets that augment VLR instances with natural language explanations. These datasets enable fine-tuning with explanation supervision to jointly learn answer prediction and rationale generation for task-specific interpretable reasoning processes. This method, as demonstrated in Figure~\ref{fig:VLR} showing that generating answer together with language explanations can improve evidence localization, have emerged as a promising direction for jointly improving grounding and knowledge integration~\cite{schwenk2022okvqa}.  However, prior work shows that such explanations often remain post-hoc and detached from the underlying reasoning process~\cite{Parcalabescu2023OnMF}.

\begin{figure*}[t]
    \centering
    \begin{subfigure}[t]{0.33\linewidth}
        \centering
        \includegraphics[width=\linewidth]{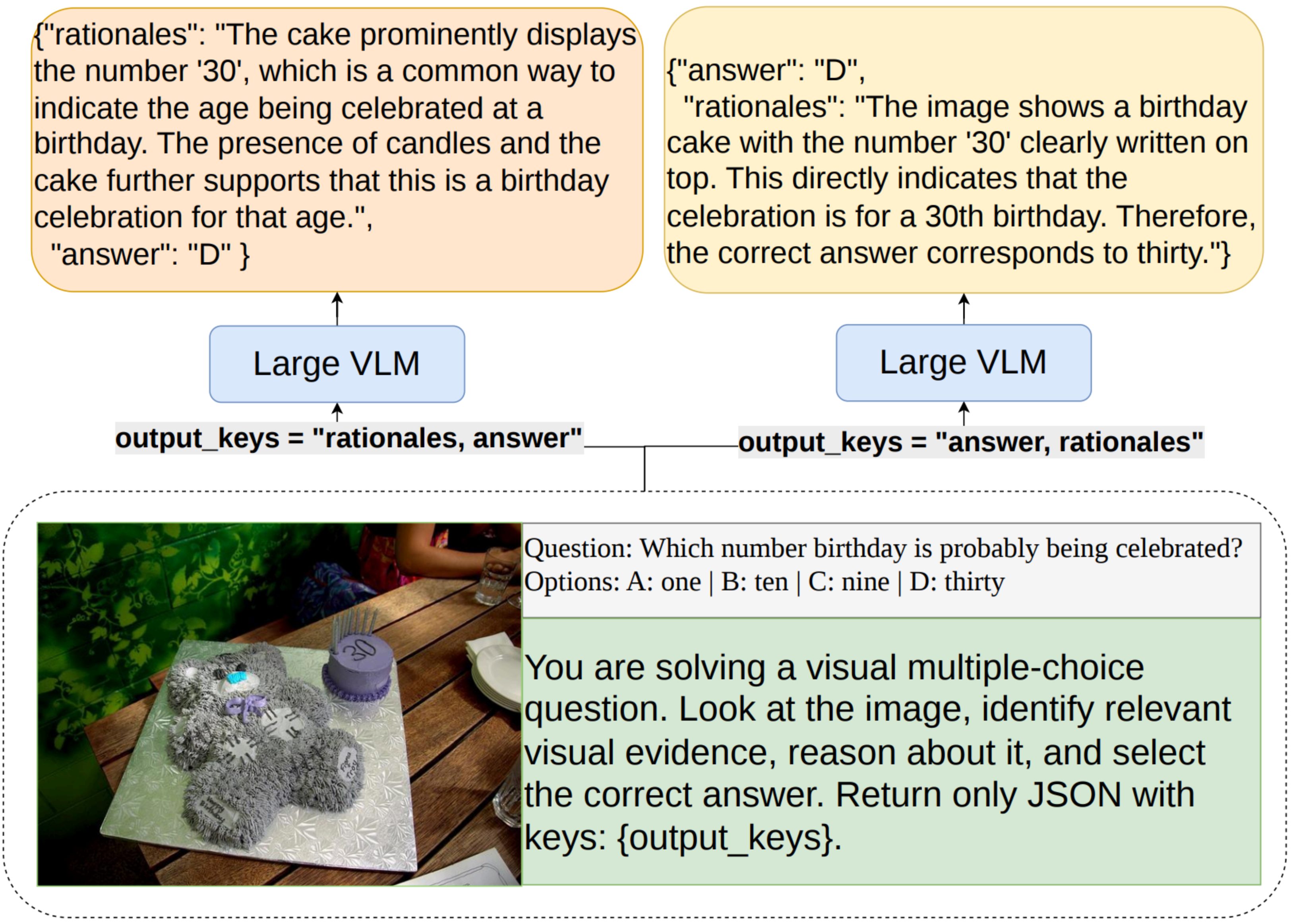}
        \caption{Illustration of one step response variations, rationale-first and answer-first, in a visual multiple-choice VQA example from the A-OKVQA validation set.}
        \label{fig:img1}
    \end{subfigure}
    \hfill
    \begin{subfigure}[t]{0.33\linewidth}
        \centering
        \includegraphics[width=\linewidth]{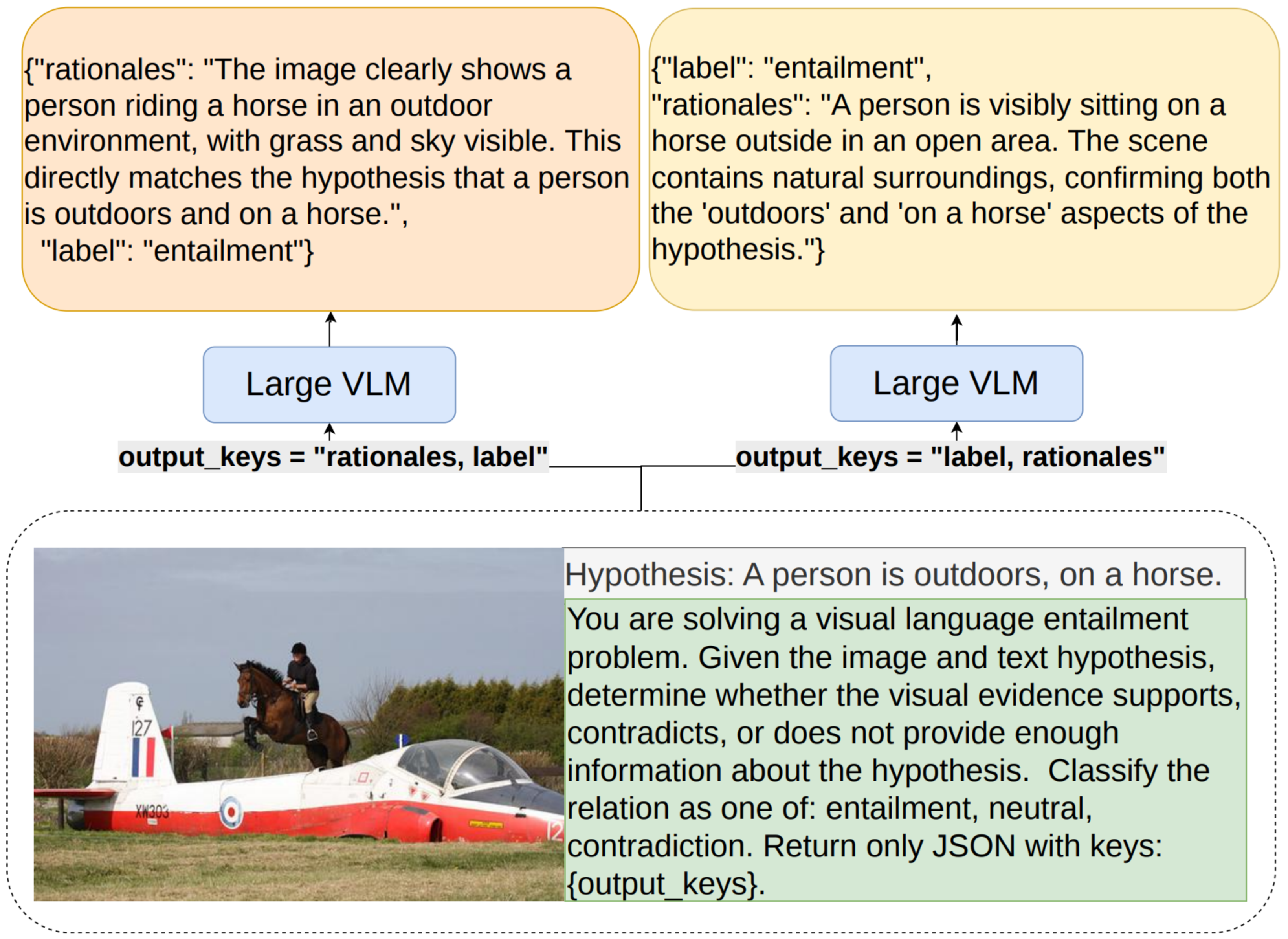}
        \caption{Illustration of an visual entailment example from the e-SNLI-VE test set. It predicts the relationship (\textit{entailment, neutral, contradiction}) between the image and hypothesis text. }
        \label{fig:img2}
    \end{subfigure}
    \hfill
    \begin{subfigure}[t]{0.33\linewidth}
        \centering
    \includegraphics[width=\linewidth]{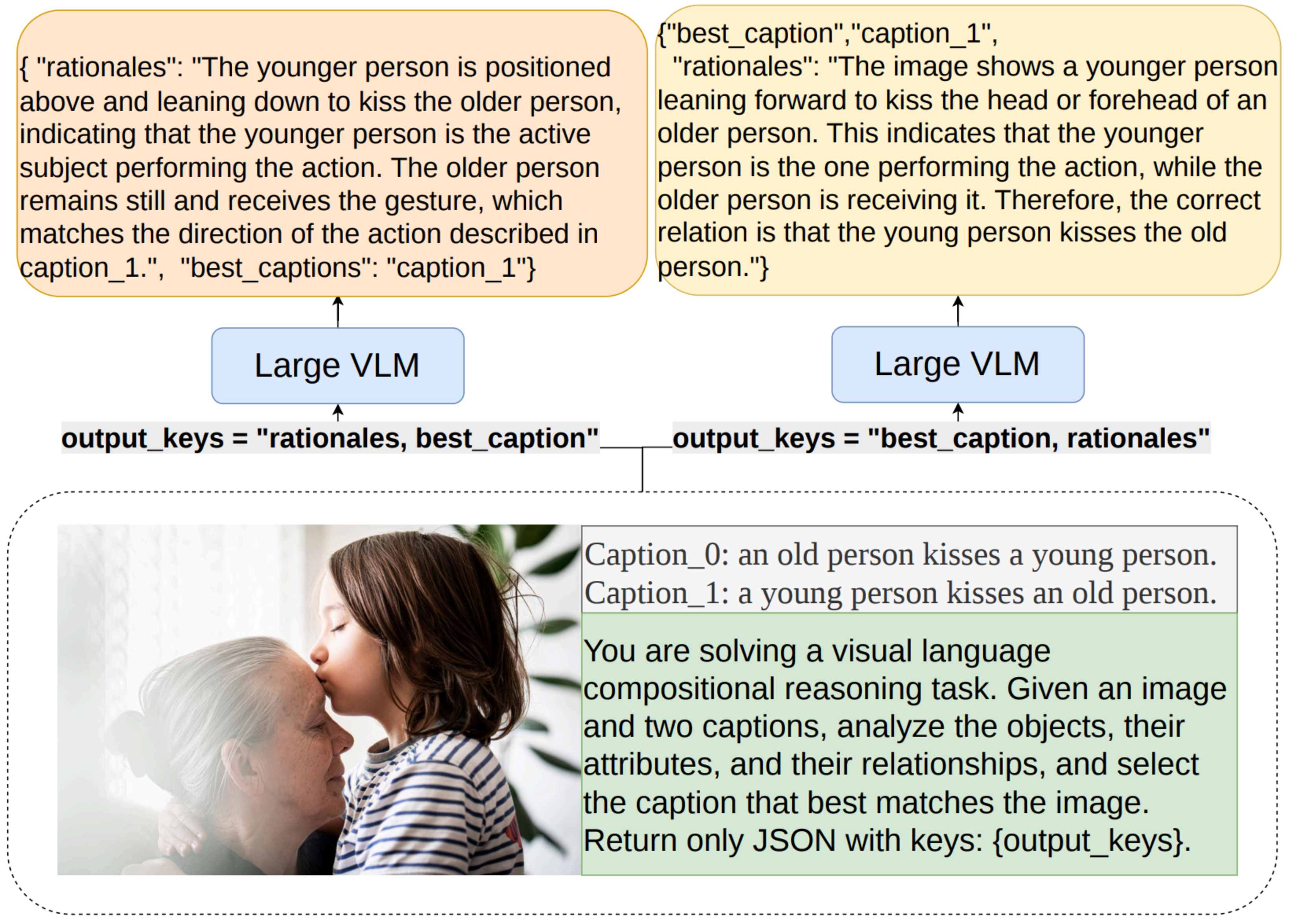}
        \caption{Illustration of  a subtask of one sample from Winoground test set. It shows a \textit{"image\_1\_vs\_captions"} subtask. }
        \label{fig:img3}
    \end{subfigure}
    \caption{The instruction prompt and output format of each VLR problem remain identical across both settings, and the only difference lies in the ordering of answer and explanation generation.}
    \label{fig:three_images}
\end{figure*}
More recently, Large Vision–Language Models (LVLMs) have demonstrated the ability to generate explanations in a zero‑shot manner through carefully designed prompting strategies, eliminating the need for explicit explanation supervisions \cite{MitraCCoT}. Chain‑of‑thought and instruction‑based prompting applied to LVLMs enables models to articulate intermediate reasoning steps that combine visual grounding with world knowledge inference \cite{zhou2024vicor}. While CoT promotes a general, task-agnostic reasoning process by decomposing inference into intermediate steps, it may introduce redundant or task-irrelevant information that can dilute focus.  

Joint answer–rationale generation tends to be more task-focused, as the rationalization is directly tied to the prediction within a single generation step, avoiding unnecessarily long or noisy reasoning chains. In this work, we systematically evaluate joint generation under different ordering strategies, comparing \textbf{rationale-first} and \textbf{answer-first} paradigms across diverse VLR tasks and models in both zero-shot and fine-tuning settings. The \textbf{rationale-first} variant preserves the efficiency of joint generation while ensuring that the final answer is explicitly conditioned on the intermediate rationale, even within a single-step process. 

Our method provides a simple yet effective diagnostic of explanation faithfulness in settings where the answer and explanation are jointly generated within a single response. If explanations truly reflect the underlying reasoning process, the generation order should have minimal impact on the predicted answer or overall output structure, and consistency scores should remain stable across settings. In contrast, significant discrepancies between answer-first and rationale-first conditions indicate that explanations are not causally grounded in the model’s decision-making process.


\section{Method}
In this work, we design a controlled experimental setup to assess whether explanations are causally aligned with model predictions within a single generation process, and to quantify their consistency across different settings. Our method consists of two key components: (i) \textbf{generation-order intervention}, where we control whether the explanation is generated before or after the answer; and (ii) \textbf{fine-tuning with explanation supervision}, where we examine how supervised explanations affect this alignment.

\subsection{Generation Order Intervention}
 To probe whether explanations are part of the reasoning process, we introduce a minimal intervention that changes the \emph{generation order} of outputs while keeping all other factors fixed. Specifically, in the \textit{rationale-first} (RF) paradigm, the model generates a short explanation before predicting an answer. In the \textit{answer-first} (AF) paradigm, the model predicts the answer directly and then produces a brief post-hoc justification. Both paradigms are implemented as prompt templates and evaluated consistently across models and datasets, as illustrated in Figure~\ref{fig:three_images}. They share a common instruction that encourages explicit reasoning grounded in visual evidence and relevant knowledge, while maintaining structured outputs (JSON) for reliable and efficient post-processing. This controlled design isolates the effect of explanation order on model behavior. While both answer‑first and rationale‑first paradigms aim to support explanation‑driven reasoning, they differ fundamentally in whether explanations merely justify predictions or actively guide them, raising the question of how explanation ordering impacts reasoning effectiveness and consistency.

\subsection{Fine-tuning with Explanation Supervision}
In the VLR task, each input $(x)$ consists of an image and its corresponding textual prompt. We construct target rationales $R^*$ post-hoc, conditioned on the ground-truth answer $A^*$. By grounding the rationale generation on $A^*$, we ensure that each explanation is consistent with the correct answer and provides a valid justification, thereby avoiding noise from potentially incorrect intermediate reasoning.

For each training sample $(x, A^*, R^*)$, we construct two alternative target sequences that differ only in the ordering of the answer and rationale: \emph{rationale-first} ($R^* \rightarrow A^*$) and \emph{answer-first} ($A^* \rightarrow R^*$). Both sequences share the same rationale $R^*$, ensuring that the supervision remains content-consistent while isolating the effect of generation order. The construction of these post-hoc rationales differs from that used in the original dataset and is described in Section~\ref{sec:experiments}.
We train the model using standard auto-regressive likelihood over both target orderings:
\begin{equation}
\mathcal{L} = -
\begin{aligned}[t]
\Big(
&\mathbb{E}_{(x, A^*, R^*)}\left[\log P(R^*, A^* \mid x)\right] \\
&+ \mathbb{E}_{(x, A^*, R^*)}\left[\log P(A^*, R^* \mid x)\right]
\Big)
\end{aligned}
\end{equation}
where each term corresponds to one supervision format. This formulation explicitly enforces that the model learns to produce both answer and explanation under different generation orders, while keeping the underlying content consistent.

\section{Experiments}
\label{sec:experiments}
\subsection{Models.} In our experiments, we select 7B‑scale models, \textbf{Qwen-2.5-VL-7B-Instruct} (Qwen-VL)\cite{bai2023qwen} and \textbf{LLaVA-1.5-7B} (LLaVA)\cite{liu2023visual_llava}, as they provide a practical balance between capacity and efficiency, enabling comprehensive experiments without requiring resource‑intensive deployment. Based on prior work and our zero-shot preliminary results, Qwen-VL demonstrates stronger reasoning capabilities compared to LLaVA. We therefore use Qwen-VL to generate rationales for our explanation-augmented fine-tuning setting. 

\subsection{Datasets.} Visual Language Reasoning (VLR) aims to jointly interpret images and text to produce grounded inferences, with challenges broadly falling into visual grounding and world knowledge inference \cite{Sammani_2023_ICCV_uni_nlx}. 
We conduct experiments on three major vision-language datasets that require different types of reasoning capabilities. Table~\ref{tab:dataset_stats} summarizes the key statistics for each dataset and split used in our experiments.

\paragraph{A-OKVQA.} This dataset is a knowledge-intensive visual question-answering benchmark. Its questions generally cannot be solved by simple knowledge base lookup and instead require integrating external knowledge grounded in the visual scene \cite{schwenk2022okvqa}.  A key feature of A-OKVQA is the inclusion of human-annotated rationales, with three free-form natural language justifications per training example, providing explicit explanations for why an answer is correct. Figure~\ref{fig:img1} illustrates an A-OKVQA example where a cake displays the number \textit{“30”}, indicating a birthday celebration. While the visual cue provides the number, correctly answering the question requires world knowledge that numbers on cakes typically denote age, allowing the model to infer that a thirtieth birthday is being celebrated. Therefore, this task requires combining visual perception with commonsense knowledge about social conventions.

The dataset contains approximately 17.1K training and 1.1K validation samples, while the test set does not provide public labels. We use the validation set for evaluation. In our experiments, we first test the quality of these human-written rationales, observing that both Qwen-VL and LLaVA models achieve near-perfect answer accuracy when conditioned on them, indicating their strong alignment with correct answers. We therefore directly leverage these rationales for supervised explanation fine-tuning. 

\paragraph{e-SNLI-VE.} Visual entailment benchmarks such as e-SNLI-VE \cite{do2020esnlive} require classifying whether an image supports a textual hypothesis as \textit{entailment}, \textit{neutral}, or \textit{contradiction}. For example, in Figure~\ref{fig:img2}, the hypothesis \textit{"A person is outdoors, on a horse."} is supported by visible evidence of a person riding a horse in an open setting, yielding \textit{entailment}. This task emphasizes fine-grained alignment between visual evidence and hypothesis semantics.

Unlike A-OKVQA, the human-written explanations in e-SNLI-VE are not designed as fine-grained rationales tightly grounded in the decision process \cite{do2020esnlive}. To obtain more controlled and answer-consistent supervision, we augment 100K training samples by generating rationales conditioned on the ground-truth labels using the Qwen-VL model. We further validate these rationales under a rationale-first setting to ensure correct label prediction, yielding high-quality triplets of image, label, and rationale, which are then used for supervised explanation fine-tuning. 

\paragraph{Winoground.} While world knowledge inference requires reasoning beyond the image using commonsense or factual knowledge, visual grounding focuses on aligning linguistic structure (entities, relations, compositional cues) with the image. Compositional benchmarks like Winoground \cite{thrush2022winoground} expose widespread failures in relational grounding in VLMs by keeping objects and words fixed while altering relations, as displayed in Figure~\ref{fig:img3}. 

This dataset is a compositional vision-language benchmark consisting of 400 test-only samples, where each sample includes two images and two contrasting captions \cite{thrush2022winoground}. The dataset is designed to evaluate fine-grained compositional reasoning, requiring models to correctly align subtle differences in visual attributes and their relationships. Each sample can be decomposed into four subtasks  as shown in Table~\ref{tab:subtasks}: two caption-to-image tasks (\textit{caption\_0\_vs\_images}, \textit{caption\_1\_vs\_images}) and two image-to-caption tasks (\textit{image\_0\_vs\_captions}, \textit{image\_1\_vs\_captions}). We split the 400 samples into 200 for training and 200 for evaluation, resulting in 800 instances for each split (4 subtasks per sample). To enable controlled fine-tuning experiments, we further generate rationales for the training instances of each subtask using Qwen-VL model, providing explanation supervision for the fine-tuning setting.
\begin{table}[t]
\centering
\small
\begin{tabular}{l l l}
\toprule
\textbf{Subtask} & \textbf{Input} & \textbf{Target} \\
\midrule
caption\_0\_vs\_images & caption 0 + image 0, image 1 & image 0 \\
caption\_1\_vs\_images & caption 1 + image 0, image 1 & image 1 \\
image\_0\_vs\_captions & image 0 + caption 0, caption 1 & caption 0 \\
image\_1\_vs\_captions & image 1 + caption 0, caption 1 & caption 1 \\
\bottomrule
\end{tabular}
\caption{Winoground subtasks in our experiments. Each original sample is decomposed into four evaluation instances covering caption-to-image and image-to-caption matching.}
\label{tab:subtasks}
\end{table}

\begin{table}[h]
\centering
\small
\begin{tabular}{llr}
\toprule
\textbf{Dataset} & \textbf{Split} & \textbf{Instances} \\
\midrule
\multirow{2}{*}{\textbf{A-OKVQA}} & Train & 17k \\
& Test & 1.1k  \\
\midrule
\multirow{2}{*}{\textbf{e-SNLI-VE}} & Train & 100k \\
& Test & 14k \\
\midrule

\multirow{2}{*}{\textbf{Winoground}} & Train & 800  \\
& Test & 800  \\

\bottomrule
\end{tabular}
\caption{Summary of dataset statistics for A-OKVQA, e-SNLI-VE, and Winoground. The number of instances corresponds to the total number of image–question–answer tuples in each dataset split.}
\label{tab:dataset_stats}
\end{table}

\subsection{Evaluation}
\paragraph{Task-Specific Metrics.}
We evaluate our vision–language models across three datasets, using task-specific metrics tailored to each setting:
\begin{itemize}
    \item For A-OKVQA, we report \textbf{accuracy} as the primary metric. A prediction is considered correct if the generated answer exactly matches the ground-truth answer. This metric is standard for visual question answering and directly reflects the model’s ability to jointly understand visual content and produce factually correct outputs.
\item For e-SNLI-VE, we compute the \textbf{F1 score} over the three entailment labels: \textit{entailment}, \textit{neutral}, and \textit{contradiction}. This metric provides a balanced evaluation of precision and recall for multi-class classification, while accounting for label imbalance. 
\item For Winoground, we also use \textbf{accuracy} to evaluate the image-caption matching. A prediction is considered correct if the model selects the correct matching image or caption for the given input. We report accuracy across all subtasks as listed in Table~\ref{tab:subtasks}.

\end{itemize}

\paragraph{Consistency Score.}
To assess explanation faithfulness, we compute a consistency score that measures the stability of model predictions across different generation orders. Let $P^{\text{AF}}$ and $P^{\text{RF}}$ denote the predictions under the answer-first and rationale-first settings, respectively. We quantify the agreement between these predictions to evaluate whether explanations remain consistently aligned with the final answers.

\begin{equation}
\text{Consistency} = \frac{1}{N} \sum_{i=1}^{N} 
\mathbf{1}\big( P^{\text{AF}}(x_i) = P^{\text{RF}}(x_i) \big)
\end{equation}

\paragraph{Overall Evaluation.}
By jointly analyzing task performance and consistency, we measure: (i) whether explanations are merely post-hoc or actively contribute to reasoning, and (ii) whether explanation supervision improves alignment between explanations and model predictions.

\section{Results}
\subsection{Performance}
\paragraph{Knowledge-intensive Visual Question Answering.}
 Table~\ref{tab:aokvqa_results_reorganized} shows  that Qwen-VL consistently outperforms LLaVA across all settings, indicating stronger reasoning capability on A-OKVQA. On the \textbf{Easy} split, both models achieve relatively high accuracy with only minor differences between rationale-first (RF) and answer-first (AF), suggesting limited sensitivity to generation order for simpler questions. In contrast, the Hard split reveals clearer order effects: Qwen-VL shows higher performance in the rationale-first setting, while LLaVA benefits more from answer-first generation. Consistency scores further support this observation: Qwen-VL maintains higher cross-order agreement than LLaVA in the zero-shot setting, indicating more stable reasoning. After fine-tuning, consistency improves substantially for both models, particularly for LLaVA. These results suggest that explanation supervision primarily enhances the alignment and robustness of predictions across generation orders on A-OKVQA dataset, rather than dramatically improving task accuracy, and that generation order becomes a more informative diagnostic under harder reasoning conditions. The results highlight that the interaction between explanation order and prediction is highly model-dependent.

\begin{table}[ht]
\centering
\resizebox{\columnwidth}{!}{%
\begin{tabular}{lccc ccc}
\toprule
 &
\multicolumn{3}{c}{\textbf{Qwen-VL}} &
\multicolumn{3}{c}{\textbf{LLaVA}} \\
\cmidrule(lr){2-4} \cmidrule(lr){5-7}
\textbf{Difficulty} & \textbf{RF} & \textbf{AF} & {\cellcolor{gray!15}\textbf{Consist.}}
& \textbf{RF} & \textbf{AF} & {\cellcolor{gray!15}\textbf{Consist.}} \\
\midrule
\multicolumn{7}{c}{\textit{zero-shot}} \\
\midrule
Easy (1075) & 86.2 & \textbf{87.8} & \cellcolor{gray!15}91.5 & \textbf{76.7} & 76.0 & \cellcolor{gray!15}82.2 \\
Hard (70)   & \textbf{80.0} & 75.7 & \cellcolor{gray!15}82.2 & 65.7 & \textbf{75.7} & \cellcolor{gray!15}74.3 \\
\midrule
\multicolumn{7}{c}{\textit{fine-tuned}} \\
\midrule
Easy (1075) & 86.5 & \textbf{88.0} & \cellcolor{gray!15}92.2 & \textbf{78.8} & 78.6 & \cellcolor{gray!15}90.5 \\
Hard (70)   & \textbf{77.1} & 75.7 & \cellcolor{gray!15}84.3 & 73.9 & \textbf{76.3} & \cellcolor{gray!15}90.0 \\
\bottomrule
\end{tabular}%
}
\caption{Answer accuracy (\%) under different response modes (RF vs.\ AF) and consistency score (\%) by task difficulty for zero-shot and fine-tuned models on the A-OKVQA validation set. Support counts are shown by difficulty category.}
\label{tab:aokvqa_results_reorganized}
\end{table}

\paragraph{Visual Language Entailment}
Table~\ref{tab:per_label_f1_metrics_restructured} shows Qwen‑VL delivers stronger, more balanced reasoning than LLaVA in all settings. Explanation order (RF vs AF) affects both models in zero-shot settings, but effects are larger and less predictable for LLaVA; for Qwen-VL the RF/AF gap is small for \textit{entailment/neutral} but large on harder \textit{contradiction} cases (79.2\% vs 52.1\%). Fine‑tuning substantially raises F1 across all labels, especially \textit{neutral} and \textit{contradiction} labels. Explanation supervision fine-tuning markedly improves explanation order consistency for both models (e.g., Qwen‑VL consistency up to 93.5\%, LLaVA up to 92.4\% for \textit{contradiction} label).

\begin{table}[ht]
\centering

\resizebox{\columnwidth}{!}{%
\begin{tabular}{lcccccc}
\toprule
 & \multicolumn{3}{c}{\textbf{Qwen-VL}} & \multicolumn{3}{c}{\textbf{LLaVA}} \\
\cmidrule(lr){2-4} \cmidrule(lr){5-7}
 \textbf{Label} & \textbf{RF} & \textbf{AF} & {\cellcolor{gray!15} \textbf{Consist.}} & \textbf{RF} & \textbf{AF} & \cellcolor{gray!15}\textbf{Consist.} \\
\midrule
\multicolumn{7}{c}{\textit{Zero-shot}} \\
\midrule
Entailment (5217) 
& 81.3 & \textbf{81.7} & \cellcolor{gray!15} 83.9 
& 63.3 & \textbf{67.2} & \cellcolor{gray!15}88.3 \\
Neutral (3799) 
& \textbf{55.1} & 54.1 & \cellcolor{gray!15} 70.3
& \textbf{14.2} & 11.4 & \cellcolor{gray!15}69.0 \\
Contradiction (5720) 
& \textbf{79.2} & 52.1 & \cellcolor{gray!15} 58.1 
& 62.4 & \textbf{75.6} & \cellcolor{gray!15}63.5 \\
\midrule
\multicolumn{7}{c}{\textit{ fine-tuned}} \\
\midrule
Entailment (5217) 
& 83.4 & \textbf{87.1} & \cellcolor{gray!15}88.2 
& 80.5 & \textbf{83.4} & \cellcolor{gray!15}89.3  \\
Neutral (3799) 
& 71.0 & \textbf{73.5} & \cellcolor{gray!15}86.3 
& 67.6 & \textbf{72.6} & \cellcolor{gray!15}75.5 \\
Contradiction (5720) 
& 87.4 & \textbf{88.4} & \cellcolor{gray!15}93.5 
& 83.5 & \textbf{86.4} & \cellcolor{gray!15}92.4 \\
\bottomrule
\end{tabular}%
}
\caption{Per-label F1 scores (\%) and consistency (\%) for e-SNLI-VE test set. Support counts are shown per label.}
\label{tab:per_label_f1_metrics_restructured}
\end{table}

\subsection{Vision Language Compositional Reasoning} 
The results in Table~\ref{tab:winoground_zs_ft_with_consistency} show two distinct failure modes. First, in zero-shot Qwen-VL, RF vs AF can swing widely (for example 71.5\% vs 42.5\% and 55.5\% vs 82.5\%), showing substantial explanation-order sensitivity. Secondly, performance of LLaVA strongly suggests a first-position bias: it is predicting the first image as best matched, resulting in near-perfect on \textit{Caption\_0\_vs\_Images} (94.7\%/95.0) but collapses on \textit{Caption\_1\_vs\_Images} (0.8\%/0.8\%), and similarly drops on \textit{Image\_1\_vs\_Captions} (8.5\%/0.7\%). The very high consistency in some of these settings indicates that predictions are stable but consistently biased. By contrast, Qwen-VL is less position-biased but more sensitive to response format. After fine-tuning, some gaps shrink, in particular for LLaVA model. While fine-tuning makes performance more balanced, compositional reasoning challenges remain for both models. A likely reason is limited supervision: only 200 training samples (800 subtask instances) are probably insufficient to reliably learn robust explanation-driven compositional reasoning.
\begin{table}[ht]
\centering

\resizebox{\columnwidth}{!}{%
\begin{tabular}{lccc ccc}
\toprule
 &
\multicolumn{3}{c}{\textbf{Qwen-VL}} &
\multicolumn{3}{c}{\textbf{LLaVA}} \\
\cmidrule(lr){2-4}\cmidrule(lr){5-7}
\textbf{Sub-Task} & \textbf{RF} & \textbf{AF} & {\cellcolor{gray!15}\textbf{Consist.}} 
& \textbf{RF} & \textbf{AF} & {\cellcolor{gray!15}\textbf{Consist.}} \\
\midrule

\multicolumn{7}{c}{\textit{zero-shot}} \\
\midrule
Caption\_0\_vs\_Images & 71.5 & 42.5 & \cellcolor{gray!15}66.5 & 94.7 & 95.0 & \cellcolor{gray!15}94.6 \\
Caption\_1\_vs\_Images & 55.5 & 82.5 & \cellcolor{gray!15}71.5 & 0.8  & 0.8  & \cellcolor{gray!15}98.7\\
Image\_0\_vs\_Captions & 84.0 & 76.5 & \cellcolor{gray!15}87.5 & 44.5 & 90.2 & \cellcolor{gray!15}50.1 \\
Image\_1\_vs\_Captions & 74.0 & 87.0 & \cellcolor{gray!15}81.5 & 8.5  & 0.7  & \cellcolor{gray!15}63.2 \\

\midrule
\multicolumn{7}{c}{\textit{fine-tuned}} \\
\midrule
Caption\_0\_vs\_Images & 66.5 & 69.5 & \cellcolor{gray!15}84.0 & 51.5 & 39.0 & \cellcolor{gray!15}59.5 \\
Caption\_1\_vs\_Images & 61.5 & 62.0 & \cellcolor{gray!15}80.5 & 50.5 & 65.0 & \cellcolor{gray!15}52.5 \\
Image\_0\_vs\_Captions & 79.0 & 90.0 & \cellcolor{gray!15}79.0 & 77.0 & 72.5 & \cellcolor{gray!15}64.5 \\
Image\_1\_vs\_Captions & 65.0 & 61.5 & \cellcolor{gray!15}69.5 & 36.0 & 39.0 & \cellcolor{gray!15}63.0 \\

\bottomrule
\end{tabular}%
}
\caption{Winoground sub-task accuracy (\%) RF and AF response modes, and consistency (\%) between RF and AF predictions, for zero-shot and fine-tuned models.}
\label{tab:winoground_zs_ft_with_consistency}
\end{table}

The results indicate that the impact of explanation order on model predictions varies significantly across models, with Qwen-VL exhibiting stronger reasoning capability in zero-shot settings, while supervised fine-tuning further enhances prediction robustness and cross-order consistency for LLaVA model.

\subsection{Error Analysis }
\paragraph{Parsing Errors.} Rationale-first (RF) generation introduces substantially more output noise and parsing ambiguity compared to answer-first (AF). In particular, Qwen-VL under RF frequently produces code‑fenced JSON blocks and free-form abstentions (e.g., \textit{“None of the options provided”} for A-OKVQA samples or generate a caption instead of simply predicting the best matched choice of \textit{"caption\_0"} on Winoground task), which deviate from the expected answer format and break simple label extractors.  While AF reduces outright abstentions, it still exhibits inconsistent formatting in explanation generation, especially with LLaVA model. It generally outputs cleaner single-letter predictions, but it remains prone to occasional long free-text responses and repetitive fragments. In zero-shot setting, rationale-first settings introduce higher parsing complexity and noise, making downstream extraction of structured answers less robust compared to answer-first generation. Fine-tuning helps enforce consistent output formatting and largely eliminates parsing errors, enabling more reliable extraction of structured predictions across both generation settings. 

\paragraph{Qualitative Analysis. }
\begin{table}[t]
\centering
\scriptsize
\setlength{\tabcolsep}{2.8pt}
\begin{tabularx}{\columnwidth}{|l|X|}
\hline
\textbf{Response Mode} & \textbf{Response} \\
\hline

\multicolumn{2}{|c|}{Qwen-VL Zero-shot} \\
\hline

Rationale-first &
{\ttfamily
\{\newline
"rationale": "The orange vehicles in the image appear to be small utility vehicles typically used by airport ground crews for transporting equipment or personnel. They are not police cars, shuttles for passengers, or part of air traffic control systems.",\newline
"answer": "B. shuttle"\newline
\}} \\
\hline

Answer-first &
{\ttfamily
\{\newline
"answer": "B. shuttle",\newline
"rationale": "The orange vehicles in the image appear to be small shuttles or service vehicles typically used on airport runways for transporting luggage, supplies, or personnel between different areas of the airport."\newline
\}} \\
\hline

\multicolumn{2}{|c|}{Qwen-VL Fine-tuned} \\
\hline

Rationale-first &
{\ttfamily
\{\newline
"rationale": "The orange vehicles are used to shuttle people around. The orange vehicles are shuttles. The orange vehicles can carry many people quickly.",\newline
"answer": "B. shuttle"\newline
\}} \\
\hline

Answer-first &
{\ttfamily
\{\newline
"answer": "B. shuttle",\newline
"rationale": "The orange vehicles shuttle around the airport. The orange vehicles move quickly to transport people or luggage. The orange vehicles are shuttles."\newline
\}} \\
\hline

\multicolumn{2}{|c|}{LLaVA Zero-shot} \\
\hline

Answer-first &
{\ttfamily
\{\newline
"answer": "D",\newline
"rationale": "The orange vehicles are air traffic controllers, ensuring the safe and efficient movement of airplanes on the runway."\newline
\}} \\
\hline

Rationale-first &
{\ttfamily
\{\newline
"rationale": "The orange vehicles are likely shuttles or trucks that are used to transport passengers or luggage to and from the airplanes. They are present near the airplanes on the runway, which suggests that they are involved in the airport operations.",\newline
"answer": "B"\newline
\}} \\
\hline

\multicolumn{2}{|c|}{LLaVA Fine-tuned} \\
\hline

Rationale-first &
{\ttfamily
\{\newline
"rationale": "The orange vehicles are used to guide the planes. The orange vehicles are used to guide the planes. The vehicles are moving around an airport, which is where air traffic control takes place. air traffic controllers use vehicles to communicate with pilots.",\newline
"answer": "D. air traffic"\newline
\}} \\
\hline

Answer-first &
{\ttfamily
\{\newline
"answer": "D. air traffic",\newline
"rationale": "The orange vehicles are used to transport passengers. The orange vehicles are shuttles. The vehicles are shuttles."\newline
\}} \\
\hline

\end{tabularx}
\caption{Response outputs from Qwen-VL and LLaVA across settings and response modes.}
\label{tab:qwen_llava_outputs_column}
\end{table}
Qwen-VL exhibits a strong and persistent prediction across all experimental settings. As the generated responses displayed in Table~\ref{tab:qwen_llava_outputs_column} regarding the example shown in Figure \ref{fig:airport_example}, neither response ordering (rationale-first vs.\ answer-first) nor fine-tuning alters its predictions. In contrast, LLaVA demonstrates substantially higher sensitivity to generation order and benefits more noticeably from fine-tuning. After fine-tuning, LLaVA achieves the correct answer in both response modes; however, its rationales remain partially inconsistent or unfaithful to the predicted answer. For example, the fine-tuned answer-first output predicts \textit{``D. air traffic''} while simultaneously describing the vehicles as passenger shuttles. While fine-tuning can improve final answer accuracy, it does not necessarily ensure coherent or semantically aligned explanations. The findings also highlight the strong influence of generation order on multimodal reasoning behavior, particularly for LLaVA, where rationale generation order can directly affect prediction outcomes.

\begin{figure}[h]
\centering
\includegraphics[width=\columnwidth]{images/aokvqa_26.png}
\caption{\textbf{Question:} What are the orange vehicles for? \\
\textbf{Options:} A. police \quad B. shuttle \quad C. passengers \quad D. air traffic \\
\textbf{Gold Answer:} D. air traffic\\
\textbf{Rationales:}
\textit{
"There are large planes.",
"They are there for the pilots to tell them where to go or not go.",
"They keep an eye on the planes."
}}
\label{fig:airport_example}
\end{figure}

\section{Related Work}
Recent advances in vision–language reasoning (VLR) have increasingly focused on integrating natural language explanations (NLEs) into prediction pipelines. Early efforts, such as e-ViL \cite{Kayser_2021_ICCV_eug}, established a benchmark for explainable vision–language tasks and evaluated models that generate human-readable explanations alongside answers. These approaches emphasize explanation quality and interpretability, but primarily treat explanations as auxiliary outputs rather than tightly coupled reasoning processes.  NLX-GPT \cite{Sammani_2022_CVPR_nlx_gpt} enables simultaneous prediction and explanation generation also through explanation-supervision fine-tuning for VLR . However, the explanation generation process is often only loosely aligned with the internal reasoning that produces the answer, and explanations may remain post-hoc in nature.

More recently, reasoning-oriented methods have explored intermediate reasoning steps, such as chain-of-thought (CoT) prompting and compositional reasoning approaches applied to challenging benchmarks like Winoground. These methods aim to improve performance by explicitly generating step-by-step reasoning. However, empirical gains on compositional vision–language tasks remain limited, suggesting that simply adding intermediate reasoning steps does not necessarily yield substantial improvements in fine-grained alignment or compositional understanding \cite{MitraCCoT}.

Datasets such as A-OKVQA \cite{schwenk2022okvqa} and \cite{lu2022learn_scienceqa} further promote explanation-driven VLR by providing rationales grounded in commonsense and external knowledge, enabling models to learn explanation-supported reasoning. These datasets highlight the importance of multi-step inference and knowledge integration, but existing work largely focuses on improving prediction accuracy or explanation quality, without systematically analyzing how explanation order influences the reasoning process itself.

\section{Conclusion}
Despite these developments, prior work has largely overlooked a key factor: the \emph{generation order} of explanations and answers. In practice, explanations may be generated before, after, or jointly with the answer within a single response, yet the impact of this ordering on model behavior has remained underexplored. In this work, we explicitly isolate and analyze the effect of generation order under controlled settings, revealing that explanation order can significantly alter model predictions. Our results show that Qwen-VL exhibits stronger reasoning capability but remains sensitive to generation order, while LLaVA demonstrates weaker reasoning and, in some cases, strong positional biases. Moreover, we find that explanation supervision via fine-tuning substantially improves consistency and robustness across generation orders, though residual order sensitivity persists, especially on compositional tasks. These findings suggest that explanations are not merely auxiliary outputs but interact directly with the reasoning process, highlighting the need to consider generation structure when evaluating explanation faithfulness in vision–language models.




\section*{Declaration on the Use of AI Assistants}
During the preparation of this work, we used GitHub Copilot to assist with the development of the code. After using this AI tool, we carefully reviewed and validated all the generated code. We take full responsibility for the accuracy, integrity and final content of this publication.

\section*{Acknowledgments}
This work is funded by the German Federal Ministry of Education and Research (BMBF) under grant numbers 01IW23002 (No-IDLE) and 01IW24006 (NoIDLEChatGPT), and by the Endowed Chair of Applied AI at the University of Oldenburg. 
\bibliographystyle{named}
\bibliography{ijcai26}

\end{document}